%% file: ms.tex
\documentclass[twocolumn,english,letterpaper]{ieeeconf}
\usepackage[T1]{fontenc}
\usepackage[latin9]{inputenc}
\usepackage[letterpaper]{geometry}
\usepackage{color}
\usepackage{refstyle}
\usepackage{amsmath}
\usepackage{graphicx}

\makeatletter

\AtBeginDocument{\providecommand\figref[1]{\ref{fig:#1}}}
\AtBeginDocument{\providecommand\subsecref[1]{\ref{subsec:#1}}}
\AtBeginDocument{\providecommand\Figref[1]{\ref{Fig:#1}}}
\RS@ifundefined{subsecref}
  {\newref{subsec}{name = \RSsectxt}}
  {}
\RS@ifundefined{thmref}
  {\def\RSthmtxt{theorem~}\newref{thm}{name = \RSthmtxt}}
  {}
\RS@ifundefined{lemref}
  {\def\RSlemtxt{lemma~}\newref{lem}{name = \RSlemtxt}}
  {}

\newenvironment{lyxcode}
	{\par\begin{list}{}{
		\setlength{\rightmargin}{\leftmargin}
		\setlength{\listparindent}{0pt}
		\raggedright
		\setlength{\itemsep}{0pt}
		\setlength{\parsep}{0pt}
		\normalfont\ttfamily}%
	 \item[]}
	{\end{list}}

\usepackage{stfloats}
\usepackage{cite}

\makeatother

\usepackage{babel}
\begin{document}

\title{Design and Take-Off Flight of a Samara-Inspired Revolving-Wing Robot}

\author{Songnan Bai and Pakpong Chirarattananon\thanks{This work was substantially supported by the Research Grants Council of the Hong Kong Special Administrative Region of China (grant number CityU-11207718).}%
\thanks{The authors are with the Department of Biomedical Engineering, City University of Hong Kong, Hong Kong SAR, China (emails:  songnabai2-c@my.cityu.edu.hk and pakpong.c@cityu.edu.hk).}
}
\maketitle
\begin{abstract}
Motivated by a winged seed, which takes advantage of a wing with high
angles of attack and its associated leading-edge vortex to boost lift,
we propose a powered 13.8-gram aerial robot with the maximum take-off
weight of 310 mN (31.6 gram) or thrust-to-weight ratio of 2.3. The
robot, consisting of two airfoils and two horizontally directed motor-driven
propellers, revolves around its vertical axis to hover. To amplify
the thrust production while retaining a minimal weight, we develop
an optimization framework for the robot and airfoil geometries. The
analysis integrates quasi-steady aerodynamic models for the airfoils
and the propellers with the motor model. We fabricated the robots
according to the optimized design. The prototypes are experimentally
tested. The revolving-wing robot produces approximately 50\% higher
lift compared to conventional multirotor designs. Finally, an uncontrolled
hovering flight is presented.
\end{abstract}
\input{sec_intro.tex}

\input{sec_modeling.tex}\input{sec_optimization.tex}

\input{sec_exp.tex}\input{sec_conclusion.tex}

\bibliographystyle{IEEEtran}
\bibliography{ms}

\end{document}

%% file: sec_intro.tex
\section{Introduction\label{sec:introduction}}

Micro Aerial Vehicles (MAVs) have gained an increasing popularity
thanks to their countless applications, such as agricultural, inspection,
and reconnaissance. With an ability to hover and the exceptional maneuverability,
rotary-wing vehicles, in particular, prove to be a highly versatile
platform. At small scales, however, flying robots suffer from limited
flight endurance due to the increased dominance of viscous forces
\cite{chen2017biologically}. Compared to the fixed-wing counterparts,
rotorcraft are less efficient. The absence of large aerodynamic surfaces
in rotary-wing designs comes with an inevitable expense of the energetic
efficiency.

To date, nature has provided us solutions for recreation of flight
with man-made machines. Bird-like morphing wings manifest high performance
aerodynamic surfaces \cite{jenett2017digital}. Taking after insects
and hummingbirds, millimeter-scale flying robots leverage unsteady
force production and leading-edge vortices as a lift enhancement mechanism
through the flapping-wing motion \cite{chen2017biologically,li2018simplified}.
In this work, we take an inspiration from winged achenes or autorotating
seeds and propose a revolving-wing robot that is capable of hovering
with a promising aerodynamic performance.

\begin{figure}
\begin{centering}
\includegraphics[width=7cm]{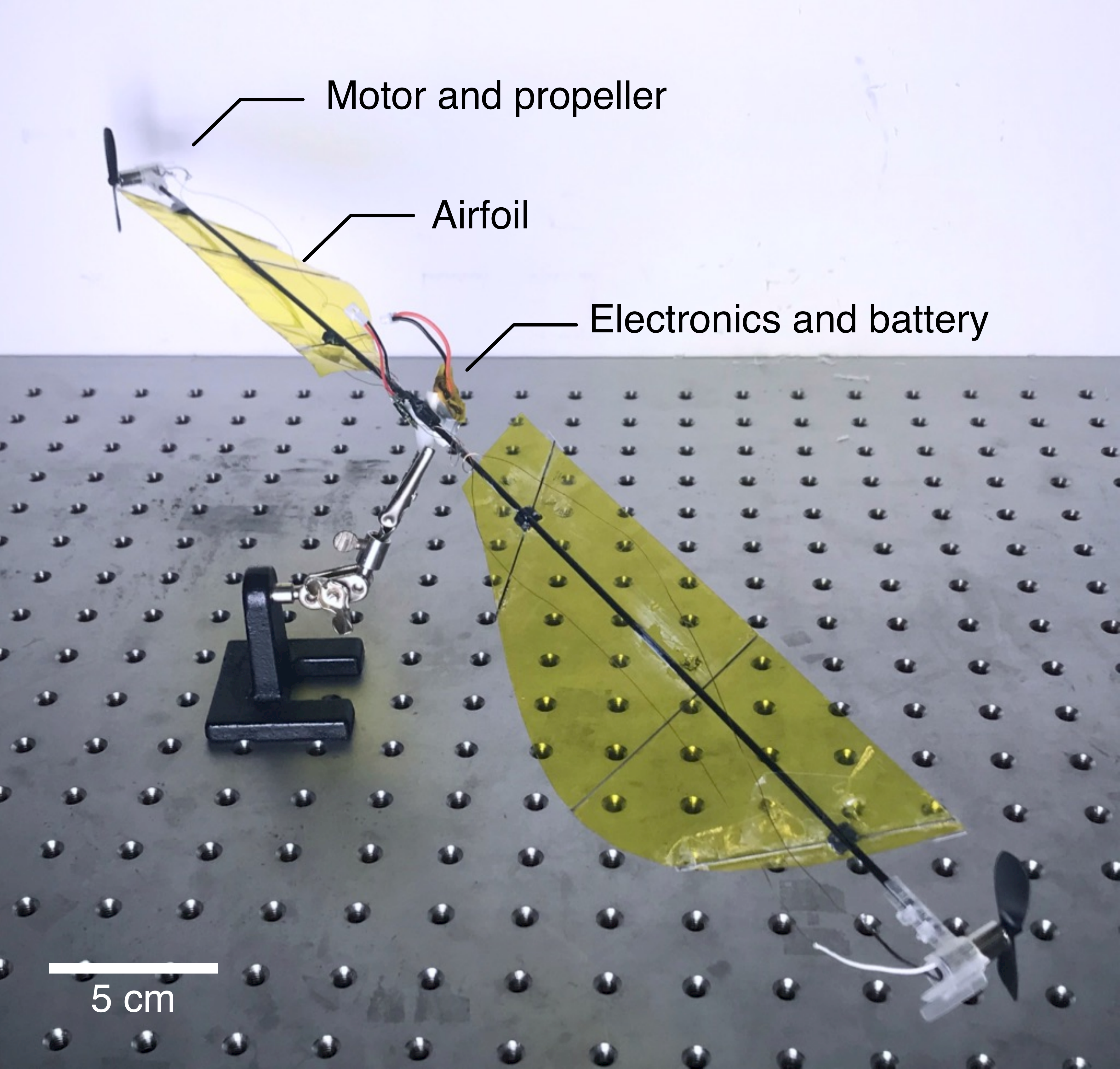}
\par\end{centering}
\caption{Photograph of a flight-capable samara-inspired robot with two large
aerodynamic surfaces. The mass of the motor-driven robot is 13.8 gram.}
\label{fig:robot_prototype}\vspace{-4mm}
\end{figure}

Similar to insect wings, a samara operates at low Reynolds numbers
and high angles of attack in comparison to conventional aircraft wings
or propeller blades. At small Reynolds numbers, large wing pitch angles
result in an improved aerodynamics performance \cite{zhao2019experimental}.
During descent, a maple seed autorotates into a helical fall and exhibits
elevated lift as a result of a stably attached leading-edge vortex
\cite{lentink2009leading,varshney2011kinematics}. Herein, we translate
these principles into a powered rotating robot in \figref{robot_prototype},
exploiting two revolving flat airfoils with high angles of attack
for thrust generation.

Unlike monocopters \cite{zhang2016controllable}, which rely solely
on propeller's thrust to stay aloft, robotic samaras resort to rotating
airfoils, relegating the role of propellers such that they are predominantly
for countering the aerodynamic drag produced by the airfoils. This
is akin to the function of propellers on fixed-wing aircraft. In \cite{ulrich2010falling},
powered flights of two robotic samaras with masses of 75 g and 38
g have been demonstrated. The robots incorporate a monowing design
that takes after natural samaras. Due to the sophisticated aerodynamics
and flight trajectories, the authors carried out over 100 iterations
to reach the two final designs that have the lift-to-weight ratios
of 1.67 and 1.36. Another related robot is presented as a 532-gram
hybrid robot with a 105-cm wingspan \cite{low2017design}. 

Unlike previous revolving robots, our proposed samara-inspired robot
adopts a symmetric dual-wing design with horizontally directed propellers.
The symmetric design simplifies the modeling efforts, resulting in
more tractable dynamic and aerodynamic models. In this configuration,
the vertical thrust is solely contributed by two wings. To create
a lightweight vehicle that hovers efficiently, we borrow fabrication
and aerodynamic modeling techniques from both flapping-wing robots
\cite{chen2017biologically,li2018simplified} and rotary-wing aircraft
\cite{bangura2016aerodynamics}. Employing momentum theory (MT) and
blade element method (BEM), the aerodynamics of airfoils are modeled
according to the wing profile \cite{zhao2019geometry}. To further
promote payload capacity and the thrust-to-weight ratio, we take into
account the actuators' dynamics. Since the performance of the motors
and propellers on the robot is affected by the revolving motion because
the translating motion of the propeller, which substantially depends
on the revolving rate of the robot, influences the propeller's thrust,
this impacts the ability of the propellers to overcome the aerodynamic
drag produced by the airfoils. By taking into account the aerodynamics
of the airfoils, the propellers, and steady-state motor model simultaneously,
we present a framework for determining the optimal geometry of the
robot. The performance of the fabricated robots is empirically examined
against the model predictions. Finally, we demonstrate a passively-stable
open-loop flight of the robot with the optimal design.\vspace{-1mm}

%% file: sec_modeling.tex
\section{Robot Dynamics}

\subsection{Overview}

\begin{figure*}
\begin{centering}
\includegraphics[width=17cm]{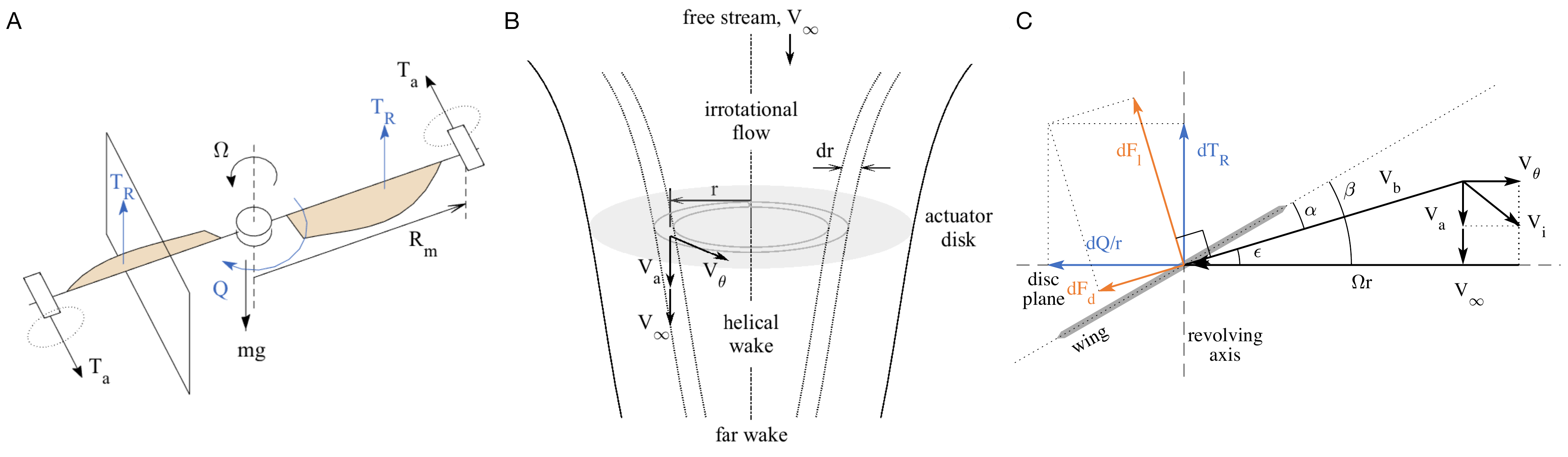}
\par\end{centering}
\caption{(A) A free body diagram illustrating the dynamics of the robot in
a revolving flight. (B) Streamtube and annular element for the momentum
theory. The induced velocity is presented as the axial and tangential
velocities. (C) Forces and relative airflow velocities experienced
by a wing element as seen from a cross section shown in (A).\label{fig:robot_dynamics_panel}}
\vspace{-4mm}
\end{figure*}

The proposed robot, schematically depicted in \figref{robot_dynamics_panel}A,
takes an inspiration from a winged seed or a samara. Vertical thrust
is produced by two large, revolving aerodynamic surfaces---airfoils.
Two motor-driven propellers are horizontally aligned in the opposite
directions. The propellers' thrusts make no direct contribution to
the robot's vertical thrust, and are solely for producing a rotational
torque in order to counter the torque created by the aerodynamic drag
from the revolving wings.

\textcolor{black}{To simplify the analysis, we consider first the
dynamics of the airfoils and then, the dynamics of the motor-driven
propellers, neglecting the possible flow interaction. In }\subsecref{AirfoilModel-1},
we employ classical momentum theory (MT) and blade element method
(BEM) to demonstrate that when the robot revolves at the angular rate
$\Omega$, the thrust generated by each airfoil ($T_{R}$), and the
associated drag torque ($Q$) about the yaw axis of the robot follows

\begin{equation}
T_{R}=C_{T,R}\Omega^{2}\text{\ensuremath{\quad}and\ensuremath{\quad}}Q=C_{Q}\Omega^{2},\label{eq:CLWCQW}
\end{equation}
where $C_{T,R}$ and $C_{Q}$ are the corresponding thrust and torque
coefficients that can be evaluated according to the geometry of the
airfoils.

For actuation, the complication arises as the propellers, placed at
distance $R_{m}$ from the revolving axis (see \figref{robot_dynamics_panel}A),
have non-zero translational velocities. In \subsecref{ActuatorModel-1},
we show that the thrust generated by each motor-driven propeller ($T_{a}$)
depends not only on its driving voltage ($U$), but also on its axial
speed ($\Omega R_{m}$). In other words, the dynamics of the propellers
are influenced by the robot's revolving rate. In equilibrium, or the
hovering condition, the yaw torque resulted from the aerodynamic drag
from the airfoils must be balanced by the yaw torque contributed by
the thrusts generated by the propellers: $Q=T_{a}R_{m}$.\vspace{-2mm}

\subsection{\textcolor{black}{Airfoil Aerodynamics\label{subsec:AirfoilModel-1}}}

To analyze the aerodynamics of the airfoils, we employ MT and BEM.
MT assumes the conservation of mass and momentum to relate the thrust
and aerodynamic drag torque of the propeller with the induced air
velocity. On the other hand, BEM exploits the quasi-steady assumption
to provide the aerodynamic forces as functions of the local air speed.
Together, the aerodynamic forces are evaluated according to the revolving
speed of the airfoil and its geometry.

\subsubsection{Momentum theory}

Momentum theory regards spinning propellers or revolving airfoils
as an infinitesimally thin actuator disc. The airflow is confined
within the boundaries of an axisymmetric streamtube as shown in \figref{robot_dynamics_panel}B.
The conservation of mass imposes a constant flow rate throughout.
The downwards free-steam velocity ($V_{\infty}$) represents the upstream
air velocity distant from the disc. The actuator disc produces the
thrust in the form of a pressure discontinuity  immediately above
and below the disc. Simultaneously, it introduces the additional air
velocity, known as the induced velocity $\mathbf{V}_{i}\left(r\right)$,
to the wake. This induced velocity is assumed to vary as a function
of the radial position ($r$). Due to the rotational motion of the
airfoils, the induced velocity is a vector sum of the axial flow ($V_{a}\left(r\right)$)
and the tangential flow ($V_{\theta}\left(r\right)$).

To compute the thrust generated by the actuator disc, one considers
an annular element of the flow with the radius $r$ and width ${\rm d}r$.
As given in \cite{macneill2017blade,rwigema2010propeller}, the elemental
thrust ${\rm d}T_{R}$ is obtained using the conservation of momentum
along the axial direction as
\begin{equation}
{\rm d}T_{R}\left(r\right)=4\pi r\rho\left(V_{\infty}+V_{a}\left(r\right)\right)V_{a}\left(r\right){\rm d}r,\label{eq:MTOriginT}
\end{equation}
where $\rho=1.2$ kg.m$^{-3}$ is the air density. Likewise, the conservation
of angular momentum relates the tangential induced velocity $V_{\theta}$
to the elemental torque ${\rm d}Q$ on the airfoils\cite{macneill2017blade,rwigema2010propeller}:
\begin{equation}
{\rm d}Q\left(r\right)=4\pi r^{2}\rho\left(V_{\infty}+V_{a}\left(r\right)\right)V_{\theta}\left(r\right){\rm d}r.\label{eq:MToriginQ}
\end{equation}
If we restrict our analysis to a hovering robot, the free-steam velocity
associated to the airfoils is zero. Equations (\ref{eq:MTOriginT})
and (\ref{eq:MToriginQ}) reduce to
\begin{align}
{\rm d}T_{R}\left(r\right) & =4\pi r\rho V_{a}^{2}\left(r\right){\rm d}r,\label{eq:MTThrust}\\
{\rm d}Q\left(r\right) & =4\pi r^{2}\rho V_{\theta}\left(r\right)V_{a}\left(r\right){\rm d}r.\label{eq:MMTTorque}
\end{align}
That is, MT provides the expressions of robot's thrust and the drag
torque in terms of the induced velocity.

\subsubsection{Blade element method\label{subsec:BladeElementMethod}}

Blade element theory considers aerodynamics forces based on the airfoil's
geometry. For a flat wing, the geometry of an airfoil is specified
by the wing pitch angle ($\beta$) the chord function ($c\left(r\right)$).
BEM radially divides the airfoil into multiple elements described
by $r$ and ${\rm d}r$. Total aerodynamic force or torque is evaluated
as a contribution of all elements.

Consider a hovering robot, its wing element at the radial position
$r$ has a translational velocity $\Omega r$ relative to the inertial
frame. According to MT above, the local airspeed is given by the induced
velocity ($\mathbf{V}_{i}$). As a result, the respective wing element
experiences the perceived airspeed ($V_{b}$) of
\begin{equation}
V_{b}=\sqrt{\left(\Omega r-V_{\theta}\right)^{2}+V_{a}^{2}},\label{eq:VB-1}
\end{equation}
as illustrated in \figref{robot_dynamics_panel}C. The BEM states
that the lift $F_{l}$ and drag $F_{d}$ forces on the wing element
are proportional to aerodynamic pressure and the associated lift and
drag coefficients ($C_{l}\left(\alpha\right)$ and $C_{d}\left(\alpha\right)$),
such that ${\rm d}F_{l,d}\left(r,\alpha\right)=\frac{1}{2}\rho V_{b}^{2}\left(r\right)C_{l,d}\left(\alpha\right)c\left(r\right){\rm d}r$,
where the angle of attack ($\alpha$) describes the orientation of
the airfoil relative to the air flow. Lift (${\rm d}F_{l}$) and drag
(${\rm d}F_{d}$) act in perpendicular and parallel to $V_{b}\left(r\right)$.
To obtain the thrust and drag torque, elemental lift and drag are
projected on to the vertical and horizontal directions. This yields
\begin{align}
{\rm d}T_{R}\left(r\right)= & N\left({\rm d}F_{l}\left(r,\alpha\right)\cos\epsilon-{\rm d}F_{d}\left(r,\alpha\right)\sin\epsilon\right),\label{eq:LiftBET-1}\\
{\rm d}Q\left(r\right)= & Nr\left({\rm d}F_{l}\left(r,\alpha\right)\sin\epsilon+{\rm d}F_{d}\left(r,\alpha\right)\cos\epsilon\right),\label{eq:TorqueBET-1}
\end{align}
where $N=2$ is the number of airfoils. The angle of attack ($\alpha$)
and the blade downwash angle ($\epsilon$) can be found from \figref{robot_dynamics_panel}C
as
\[
\epsilon=\tan^{-1}\left(\frac{V_{a}}{V_{\theta}-\Omega r}\right)\text{\ensuremath{\quad}and\ensuremath{\quad}\ensuremath{\alpha}=\ensuremath{\beta-\epsilon}}.
\]

To compute the thrust and torque coefficients (equation (\ref{eq:CLWCQW}))
of the robot, we consolidate the outcomes from MT and BEM. Equations
(\ref{eq:MTThrust})-(\ref{eq:MMTTorque}) are incorporated with (\ref{eq:LiftBET-1})-(\ref{eq:TorqueBET-1})
to eliminate the axial and tangential velocities. The elemental thrust
and torque are then integrated over the whole wing to yield the total
thrust and torque. The resultant thrust and torque coefficients can
be numerically computed according to the wing chord profile ($c\left(r\right)$)
and the pitch angle ($\beta$). \vspace{-2mm}

\subsection{\textcolor{black}{Propellers' Dynamics\label{subsec:ActuatorModel-1}}}

In our design, the propellers, driven by DC motors, are responsible
for generating the torque to counter the aerodynamic drag from the
airfoils. The revolving design requires the propellers to traverse
with respect to the inertial frame in flight. As a consequence, in
addition to the driving voltage, the aerodynamic forces produced by
the propellers depend on the revolving speed of the robot. In order
to determine the propeller's thrust, both the dynamics of the propeller
and the motor must be considered together.

\subsubsection{Propeller aerodynamics\label{subsec:Propeller-Aerodynamics-1}}

Similar to the airfoils, the aerodynamics of the propellers can be
modeled with MT and BEM, with some modifications. In the hovering
condition, the translational speed of the propeller depends on its
distance from the revolving axis ($R_{m}$) and the revolving rate
as $\Omega R_{m}$ (not to be confused with the spinning rate of the
propeller, $\omega$). From MT, this constitutes an axial free stream
velocity of the air seen by the propeller. In addition, if the tangential
induced velocity is neglected as in \cite{bangura2016aerodynamics},
the axial induced velocity ($v_{i}$) is independent of the radial
position. The thrust generated by the propeller ($T_{p}$) according
to MT takes the form resembling the integration of equation (\ref{eq:MTOriginT}):
\begin{align}
T_{p} & =2\rho\pi R_{p}^{2}v_{i}\left(v_{i}+\Omega R_{m}\right),\label{eq:MMT}
\end{align}
where $R_{p}$ is the propeller's radius. Using the same framework
as the airfoils, it has been shown in \cite{bangura2016aerodynamics}
that BEM provides an expression for the thrust of an $n$-blade propeller
with radius $R_{p}$ as
\begin{equation}
T_{p}=\frac{1}{2}\rho nR_{p}^{4}\left(a_{0}-a_{1}\frac{v_{i}+\Omega R_{m}}{\omega R}\right)\omega^{2},\label{eq:propellerBETthurst}
\end{equation}
where $a_{0}$ and $a_{1}$ are dimensionless lumped coefficients
related to the propeller's pitch and chord profiles. This equation
assumes the propeller blade has constant pitch and chord, and infinite
aspect ratio \cite{bangura2016aerodynamics}. While this is not the
case for our propellers, we believe the model still captures the dominant
aerodynamic effects. Relaxation of these assumptions primarily leads
to different definitions of the lumped parameters. Under the same
assumption, the propeller's torque due to the aerodynamic drag is
given by 
\begin{equation}
\tau_{p}=\frac{1}{2}\rho nR_{p}^{5}a_{2}\omega^{2}+\left(T_{p}\frac{\kappa v_{i}+\Omega R_{m}}{\omega}\right),\label{eq:propeller_torque}
\end{equation}
where $a_{2}$ is another dimensionless lumped parameter for the blades,
and $\kappa$ is the induced power factor which accounts for the power
loss caused by wake rotation and tip loss. Unlike standard airfoils,
lumped coefficients and the power factor for propellers are commonly
empirically determined \cite{bangura2016aerodynamics}.

We solve for the induced velocity as a function of $\Omega$ and
$\omega$ by combining the result from MT (equation (\ref{eq:MMT}))
and BEM (equation (\ref{eq:propellerBETthurst})). It follows that
the propeller's torque can be numerically evaluated for each $\Omega$
and $\omega$ pair as long as the lumped coefficients are known.

\subsubsection{First-order motor model\label{subsec:First-Order-Motor-Model-1}}

In this part, we employ a first-order model to relate the motor's
driving voltage, output torque, and the rotational rate. For a motor-driven
propeller, this directly links the driving voltage ($U$) to the propeller's
torque ($\tau_{p}$) and spinning speed ($\omega$).

In steady state, where the rotational rate is constant, the voltage
law states $U=IR_{i}+k\omega,$ where, $I$ denotes the current, $R_{i}$
is the motor\textquoteright s internal resistance, and the $k\omega$
term represents the back EMF, assumed proportional to $\omega$ with
a motor constant $k$. The motor torque is given as $\tau_{p}=kI$.
Hence,
\begin{equation}
\tau_{p}=\frac{k}{R_{i}}\left(U-k\omega\right).\label{eq:motor_model}
\end{equation}
\vspace{-2mm}

\subsection{Integrated robot dynamics}

With the aerodynamics and actuator's dynamics of the propeller analyzed,
we consolidate the findings to yield an integrated model. When combined,
equations (\ref{eq:propeller_torque}) and (\ref{eq:motor_model})
enable us to determine $\omega$ as a function of $U$ and $\Omega$.
With the expression of propeller's thrust (equation (\ref{eq:propellerBETthurst})),
we get rid of $\omega$ and obtain a direct relationship between $T_{p},$
$U$, and $\Omega$ as $T_{p}=T_{p}\left(U,\Omega\right)$. Next,
in the hovering condition, the robot's yaw torque contributed by the
propellers balances out the airfoil drags (equations (\ref{eq:CLWCQW})),
such that $2R_{m}T_{p}\left(U,\Omega\right)=Q\left(\Omega\right)$.
Finally, we find the vehicle's revolving speed for a particular motors'
driving voltage. This, when substituted back into equation (\ref{eq:CLWCQW}),
we obtain a prediction of the robot's thrust given the input voltage.
The framework enables us to estimate the generated thrust from the
robot and airfoil geometries.\textcolor{black}{}\vspace{-2mm}

%% file: sec_optimization.tex
\section{\textcolor{black}{Design Optimization\label{sec:Design-Optimization}}}

In the previous section, we investigate the dynamics of the robot.
The framework enables us to predict the total thrust generated by
the robot, given the voltage supplied to the actuators. The results
depend on relevant parameters related to aerodynamics, airfoil geometry,
and motors. In this section, we identify these model parameters and
search for the optimal wing geometry for maximizing the robot's thrust
based on the set of chosen hardware.\vspace{-2mm}

\subsection{Model Parameters}

In the airfoil model, the relevant aerodynamic parameters used for
BEM are lift and drag coefficients, $C_{l}$ and $C_{d}$. Both of
them depend on the angle of attack $\alpha$. Since the airfoils of
our prototype are flat and operate at large angles of attack with
low to intermediate Reynolds numbers ($\sim2.5\times10^{4}$). The
conditions are similar to those of large insect wings \cite{dickson2004effect,lentink2009leading}.
Several quasi-steady aerodynamic studies \cite{dickson2004effect,lee2016quasi,macneill2017blade}
have established approximate forms of lift and drag coefficients for BEM
as
\begin{align}
C_{l}\left(\alpha\right)= & C_{l,1}\sin\left(2\alpha\right),\label{eq:lift_coefficient_model}\\
C_{d}\left(\alpha\right)= & C_{d,0}+C_{d,1}\left(1-\cos\left(2\alpha\right)\right),\label{eq:drag_coefficient_model}
\end{align}
 where the coefficients $C_{l,1}$, $C_{d,0}$, $C_{d,1}$ are typically
empirically determined. Equations (\ref{eq:lift_coefficient_model})
and (\ref{eq:drag_coefficient_model}) also resemble the results from
the flat plate theory used for small gliders \cite{hoburg2009system}.
Without prior experimental data, we approximate these coefficients
from related literature as $C_{l,1}=1.72$, $C_{d,0}=0.11$, and $C_{d,1}=1.94$
\cite{lentink2009rotational}.

For the motors and propellers, our prototype uses coreless DC motors
and matched propellers from Crazyflie 2.0. The motor parameters of
$R_{i}=1.58\ \Omega$ and $k=1.1$ mV$\cdot$s$\cdot$rad$^{-1}$
are taken from previous identification results in \cite{hsiao2018ceiling}.
Similarly, the associated blade coefficients and power factor have
been given in \cite{hsiao2018ceiling} as $a_{0}=0.3633$, $a_{1}=1.9960$,
$a_{2}=0.0022$, and $\kappa=1.87$. The maximum supplied voltage
is conservatively chosen to be 3.5 V.

\begin{figure}
\begin{centering}
\includegraphics[width=7.2cm]{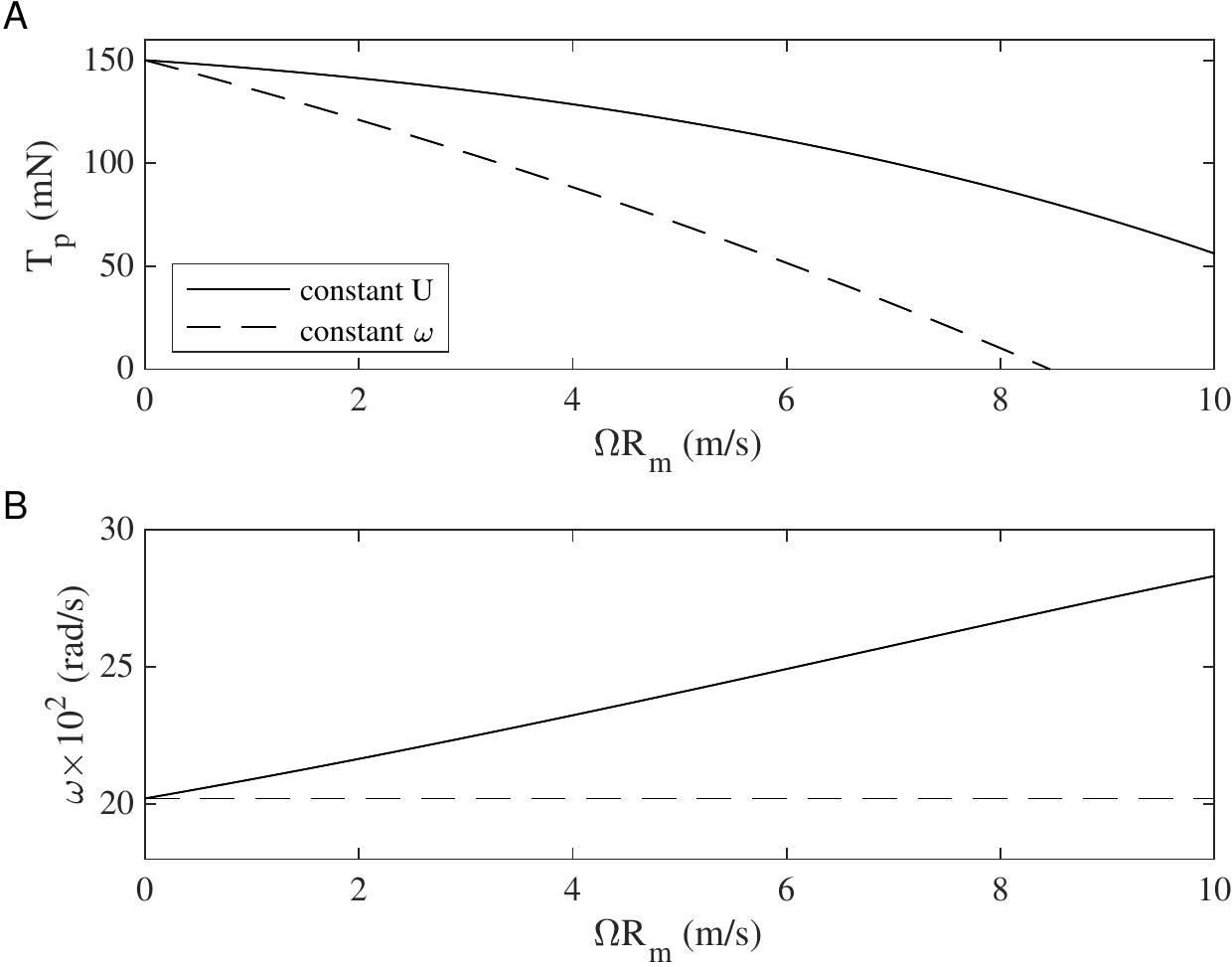}
\par\end{centering}
\caption{The propeller's thrust (A) and spinning speed (B) plotted against
the axial inflow speed. The solid lines indicate the results from
the model, assuming a constant driving voltage of 3.5 V. The dashed
lines show the thrust prediction in the scenario where $\omega$ is
held constant.. \label{fig:ActuatorsNew}}
\vspace{-4mm}
\end{figure}
With the listed parameters, \figref{ActuatorsNew} shows the model
predictions of the propeller's thrust and its rotational rate against
the linear axial speed ($\Omega R_{m}$) as outlined in Section \ref{subsec:ActuatorModel-1}.
The plots reveal that, assuming a constant voltage $U=3.5$ V, the
generated thrust drops, whereas the propeller's speed ($\omega$)
increases, as the revolving rate ($\Omega$) rises. This is because
the axial velocity effectively reduces the aerodynamic drag seen by
the propeller. For comparison, we plot the thrust prediction in case
$\omega$ remains constant. In such case, the model severely underestimates
the propeller's thrust at high axial velocity. This verifies that
it is essential to take into account the revolving rate of the robot
when analyzing the propeller's dynamics.\vspace{-2mm}

\subsection{Optimization Method}

With the proposed dynamic models and associated parameters, we proceed
to search for the optimal airfoil design and propellers' placement.
The optimization problem is setup as
\begin{equation}
{\bf x}^{*}=\max_{\mathbf{x}}T_{R}\left({\bf x}\right)-m\left({\bf x}\right)g,\label{eq:optimization}
\end{equation}
subject to some chosen constraints, where ${\bf x}\in R^{6}$ is a
set of variables to be optimized, $m\left({\bf x}\right)$ is the
mass of the robot, and $g$ is the standard gravity. The proposed
objective function, unlike the thrust-to-weight ratio, is intended
for maximizing the payload capability of the robot.  

Here, we use six design variables: wing pitch angle $\beta$, position
of the propellers $R_{m}$, wing semi-span $R_{tip}$, and three parametrized
chord lengths $c_{1}-c_{3}$. We impose the wing root position to
be at $0.15R_{tip}$. Restricting the wing's leading edge to a straight
line, the variable $c_{1}$ corresponds to the length of the chord
at the wing's root, whereas $c_{2}$ and $c_{3}$ are the chord lengths
at two intermediate locations uniformly separated between the wing's
root and the wing's tip as illustrated in \figref{ResultofDesignWing}A.
Assuming other components are unchanged, $m\left({\bf x}\right)$
is taken as the mass of the airframe and of the wings, calculated
using the linear density of the airframe (carbon fiber rod) of $4.7$
g$\cdot$m$^{-1}$ and the areal density of the wings (polyamide film
and wing spars) of $92.6$ g$\cdot$m$^{-2}$. In addition, we incorporate
the following constraints.
\begin{itemize}
\item We limit $R_{m}\geq R_{tip}$ for the ease of fabrication, such that
the motors and propellers can be physically mounted without interfering
with the wings.
\item The maximum semi-span is selected to avoid undesirably large prototypes,
such that $R_{tip}\leq23$ cm. This is because MT favors a larger
actuator disc as it requires lower power for thrust generation \cite{bangura2016aerodynamics}.
\item The chord length at the wing tip is zero. The unmodeled tip loss effects
demote the airfoil efficiency thanks to tip vortices \cite{bangura2016aerodynamics},
rendering the airfoil aerodynamically inefficient near the tip.
\end{itemize}
The final wing profile is computed from a cubic spline interpolation
of optimized variables.\vspace{-2mm}

\subsection{Optimal Airfoils}

\begin{figure}
\begin{centering}
\includegraphics[width=7.2cm]{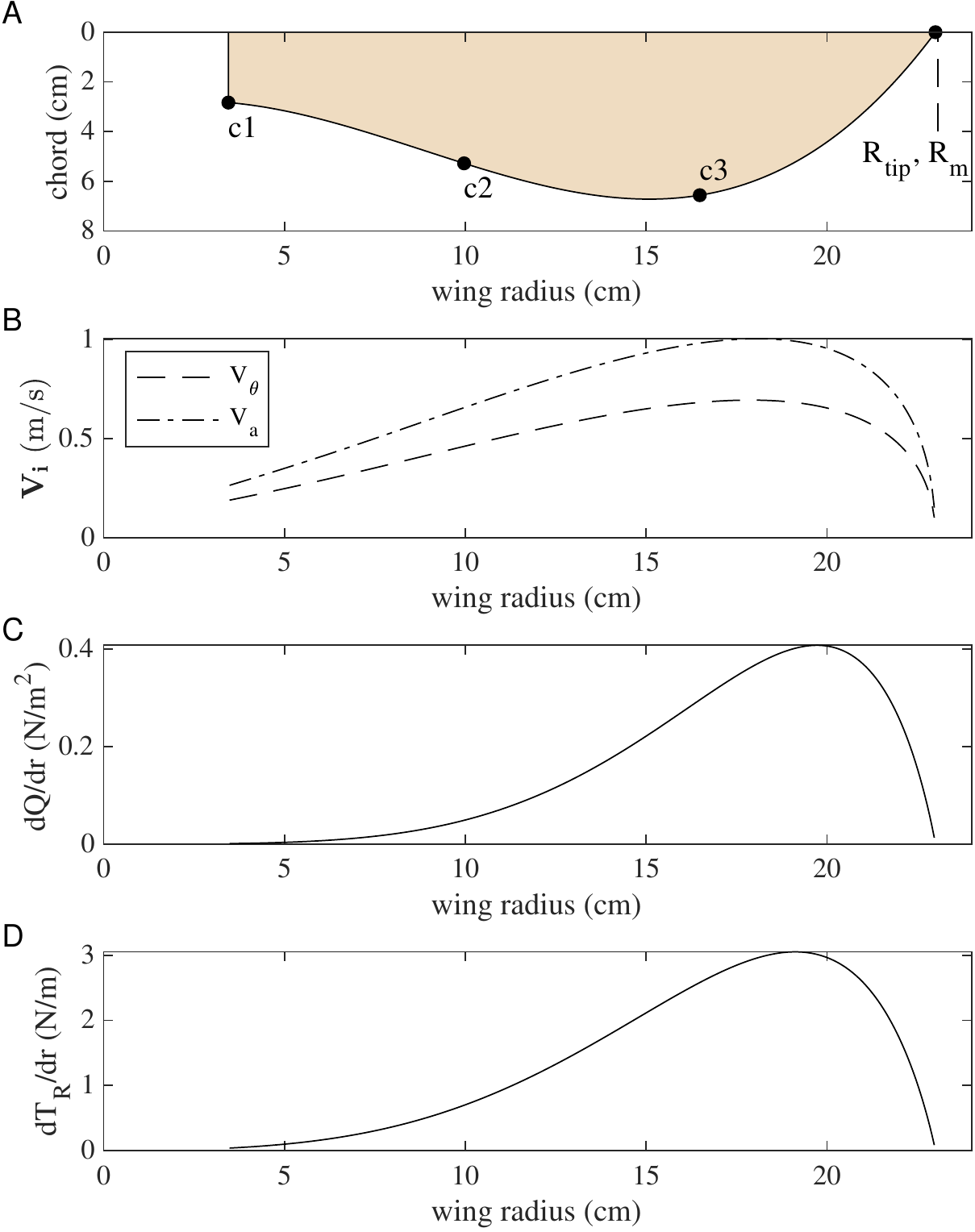}
\par\end{centering}
\caption{(A) The optimal wing geometry and location of the motor. The corresponding
wing pitch angle is $27.5^{\circ}$ (not shown). (B) The induced axial
and tangential velocities at different points along the wing, assuming
the robot revolves at the speed corresponding to $U=3.5$ V. (C) Distribution
of the drag torque. (D) Distribution of the airfoil's thrust. \label{fig:ResultofDesignWing}}
\vspace{-4mm}
\end{figure}
The proposed optimization problem is solved with the Nelder--Mead
simplex algorithm. While this does not guarantee a globally optimal
solution, we achieve a local solution with the predicted thrust of
$285$ mN. The corresponding robot profile is drawn in \Figref{ResultofDesignWing}A,
with the wing pitch angle of $27.5^{\circ}$. The optimized wing design
has a relatively narrow width near the root, with the largest chord
near the location of $c_{3}$. \Figref{ResultofDesignWing}B-D also
shows the predicted induced velocities and aerodynamic pressure at
locations along the airfoil.

 The plot of induced velocities reveal that the induced airflow is
markedly slower than the translational speed ($\sim\Omega R_{m}$),
which can be as large as $\approx5-10$ m$\cdot$s$^{-1}$. Equations
(\ref{eq:VB-1}) and (\ref{eq:LiftBET-1}), therefore, suggest that
the thrust is majorly contributed by the wing's translation. As a
consequence, the optimal wing features a wider chord further away
from the revolving axis, leading to the elemental thrust and torque
in \Figref{ResultofDesignWing}C and D that peak near the wing tip.
In the meantime, while the wing area near the root gives rise to relatively
little thrust, the chord length at the wing root is still notable.
This is because closer to the revolving axis, this area favorably
contributes less towards the drag torque.\vspace{-2mm}

\begin{lyxcode}
\end{lyxcode}

%% file: sec_exp.tex
\section{\textcolor{black}{Experimental Validation}}

\subsection{\textcolor{black}{Fabrication and Components\label{subsec:Fabrication-and-Components}}}

The robot consists of four primary components: the airframe, airfoils,
a pair of motors and propellers, and flight electronics, as shown
in \figref{robot_prototype}.

The airframe was constructed from a carbon fiber rod with 2-mm diameter.
The airfoils, or wings, were made from 250-$\mu$m polyimide film
(Kapton), laser cut to the desired geometry using CO$_{2}$ laser
(Epilog Mini 24). The wings remain flat in flight thanks to the structural
support provided by the wing spars (carbon fiber rod with 0.5-mm diameter).
The wings are attached to the airframe via small 3D printed parts
(Black resin, Form 2, Formlabs). The design of these printed parts
dictates the wing pitch angle. All parts are fixed together using
Cyanoacrylate adhesives and epoxy resin.

For actuation, we employed 23-mm radius propellers and 7$\times$16-mm
coreless DC motors commercially available as parts for Crazyflie 2.0
for the prototypes. According to the measurements in \cite{hsiao2018ceiling},
each propeller can generate $\approx90-180$ mN when the motor is
supplied by the voltage 2.5-4.0 V. The motors were driven by a micro
2.4GHz RC servo receiver with two built-in Electronic Speed Controllers
(Deltang DT Rx31d). A single-cell 100-mAh Li-ion battery was for power.
The mass of the robot is $13.8$ g.

For validation of the optimized design, we fabricated four robots
for comparison, robot \textcircled{\footnotesize{A}} with the optimal
wing geometry and wing pitch angle ($27.5^{\circ}$); robot \textcircled{\footnotesize{B}}
with the same wing geometry, but with the wing pitch angle of $11.0^{\circ}$;
robot \textcircled{\footnotesize{C}} with the same wing geometry,
but with the wing pitch angle of $40.0^{\circ}$; and robot \textcircled{\footnotesize{D}}
with the same planform area and winspan, but with a constant chord
and the wing pitch angle of $21.0^{\circ}$. All robots weigh approximately
the same as they all carry identical parts and wing areas.\vspace{-2mm}

\subsection{\textcolor{black}{Force Measurements}}

\subsubsection{Experimental Setup}

\begin{figure}
\begin{centering}
\includegraphics[width=5.6cm]{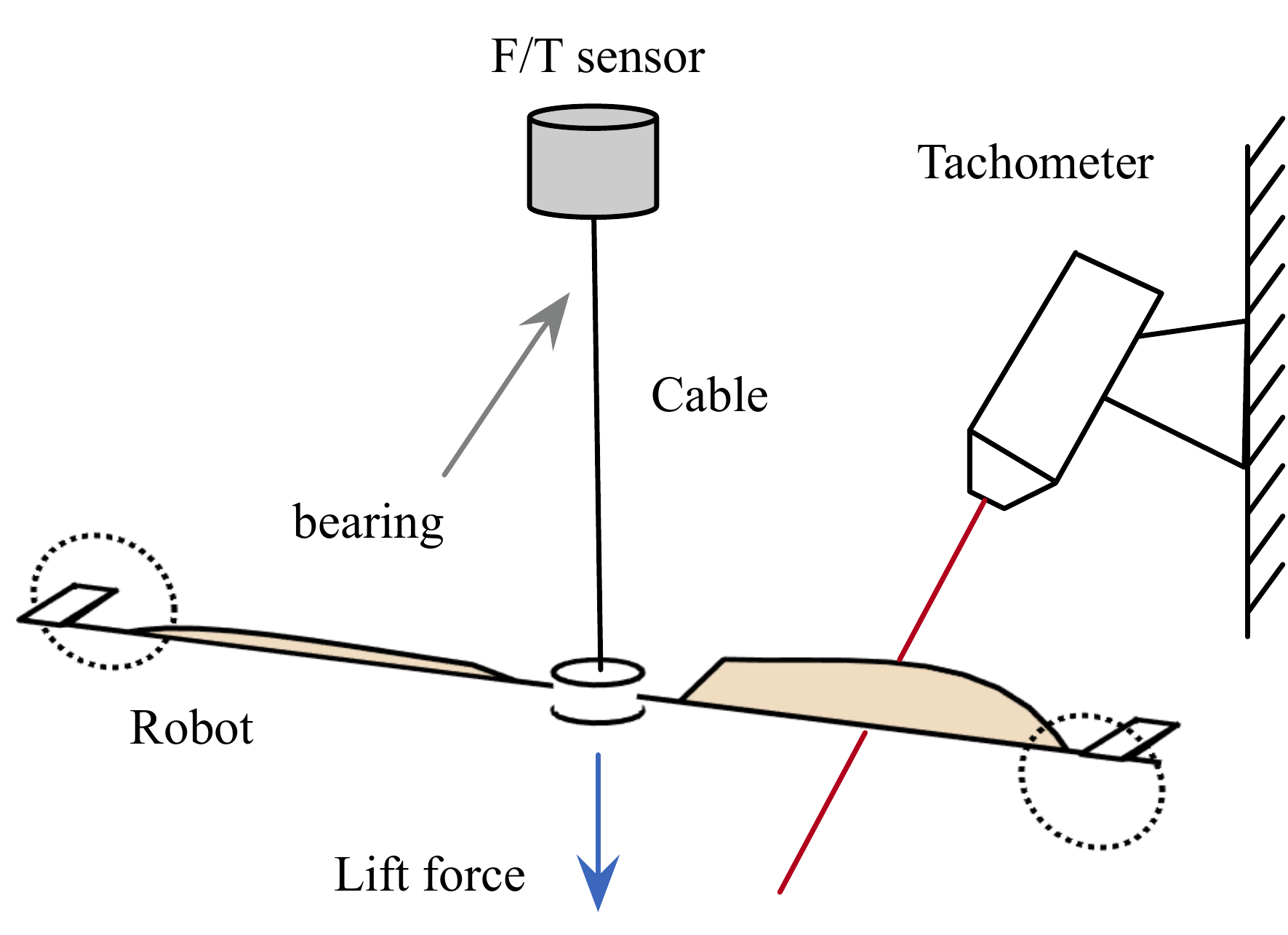}
\par\end{centering}
\caption{Experimental setup for measurements of the revolving speed and generated
thrust. \label{fig:PC}}
\vspace{-4mm}
\end{figure}

To evaluate the thrust generated by the robots, we perform static
experiments to measure the generated force when the robots were given
various commands. The experimental setup is schematically presented
in \figref{PC}. The robot was suspended upside down from the load
cell (Nano17, ATI) by a cable. The rated sensor resolution of 3.1
mN is notably lower than the expected thrust. A rolling bearing was
incorporated proximal to the load cell to allow the robot and the
cable to freely rotate. The load cell was covered for thermal insulation
to eliminate a possible convective effects due to the wake generated
by the robot. A tachometer (Advent A2108) was mounted above the robot
for measuring the revolving speed.

In this configuration, the generated thrust force directs downward
and can be measured by the sensor through the cable. Thanks to the
symmetry, during operation, the robot retained an approximate upright
orientation. The thrust is assumed vertical, or aligned with the sensor's
primary axis. The weight of the robot and other parts can be directly
subtracted from the measurements to obtain the thrust.

\subsubsection{Measurement Results\label{subsec:Measurement-Results}}

\begin{figure}
\begin{centering}
\includegraphics[width=7.2cm]{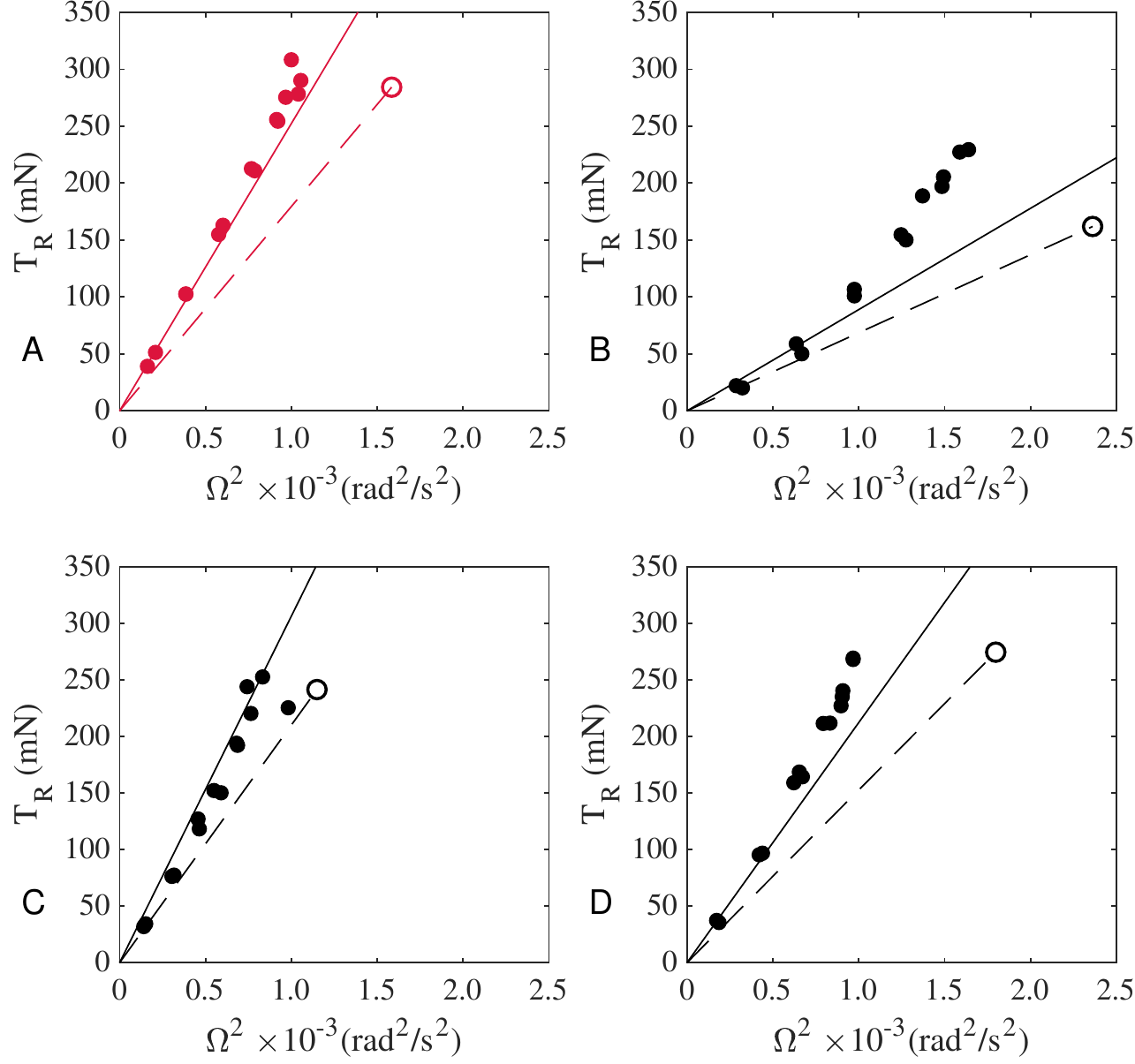}
\par\end{centering}
\caption{The thrust and revolving speed measurements from four prototypes (dots).
The dashed lines show the model predictions based on the used coefficients,
with hollow circular points representing the expected maximum thrust
when the motors are driven at $U=3.5$ V. The solid lines indicate
the refitted model predictions with the revised lift and drag coefficients.
(A) corresponds to the robot the optimal airfoils and the optimal
wing pitch angle of $27.5^{\circ}$; (B) with the wing pitch angle
of $11^{\circ}$; (C) with the wing pitch angle of $40^{\circ}$;
and (D) with the rectangular wings and the wing pitch angle of $21^{\circ}$.
\label{fig:MeasureResults}}
\vspace{-4mm}
\end{figure}
 We performed the experiments on four robots with different configurations
as presented in Section (\ref{subsec:Fabrication-and-Components}).
For each trial, we recorded the thrust, averaged from 10 s, when the
revolving speed was approximately constant. We remotely commanded
seven different voltage setpoints. Both motors on the robot received
an identical voltage. Two measurements were obtained for each setpoint.
In total, we acquired 14 datapoints for each robot, or $\text{56}$
datapoints among four robots. In all datapoints, the cycle averaged
horizontal force is less than 5\% of the vertical component.

Due to the lack of onboard measurements, we are unable to record the
actual voltage of the motors. \Figref{MeasureResults} demonstrates
the measured thrusts against the squared revolving speed. Apart from
robot \textcircled{\footnotesize{B}} (\figref{MeasureResults}B),
which has relatively low wing pitch angle, the outcomes follow an
approximately linear trend as predicted by equation (\ref{eq:CLWCQW}).

The measurements, as anticipated, verify that robot \textcircled{\footnotesize{A}},
with the optimized design, generates the thrust up to $310$ mN, equating
to the thrust-to-weight ratio of 2.3 (\Figref{MeasureResults}A).
This number markedly reduces to $\approx250$ mN for robots \textcircled{\footnotesize{B}}
and \textcircled{\footnotesize{C}}. Robot \textcircled{\footnotesize{D}},
on the other hand, has the maximum thrust of 270 mN , approximately
10\% lower than that of robot \textcircled{\footnotesize{A}}. For
comparison, we separately measured the maximum thrust force generated
by a vertically-aligned motor-propeller pair with no revolving wings
using the same driving board and battery. The force produced by two
propellers is 207 mN, significantly lower than all four revolving-wing
robot prototypes. In particular, with respect to robot \textcircled{\footnotesize{A}},
the revolving-wing configuration amplifies the lift by approximately
50\% (from 207 to 310 mN). The results highlight the benefit of the
proposed samara-inspired design.

\Figref{MeasureResults} also shows the thrust predictions according
to equation (\ref{eq:CLWCQW}) based on the parameters given in Section
(\ref{subsec:Measurement-Results}) in dashed lines. The unfilled
circular points indicate the predicted thrust when $U\approx3.5$
V. While the general trends are in agreement with the experimental
data, the model consistently underpredicts the generated thrusts and
overpredicts the revolving speeds. We believe these are primarily
due to the inaccuracy of both aerodynamic parameters (such as $C_{l}\left(\alpha\right)$
and $C_{d}\left(\alpha\right)$, $a_{i}'s$, $\kappa$, etc.) and
motor parameters (including $R_{i}$ and $k$), owing to the lack
of prior experimental data.

In an attempt to ameliorate the model's accuracy. We revise the aerodynamic
parameters $C_{l,1}$, $C_{d,0}$ and $C_{d,1}$ by finding a new
set of parameters that minimize the root mean square errors of the
thrust predictions from equation (\ref{eq:CLWCQW}) using all 56 data
points. In \figref{MeasureResults}, the solid lines present the re-fitted
thrust predictions with $C_{l,1}=2.67$, $C_{d,0}=0.22$ and $C_{d,1}=2.58$.
Without direct measurements of $U$, $T_{p}$ or $Q$, we are unable
to revise other model parameters. This leaves room for improvement
in the future.\vspace{-2mm}

\subsection{\textcolor{black}{Lift-off Flight}}

\begin{figure*}
\begin{centering}
\includegraphics[width=16cm]{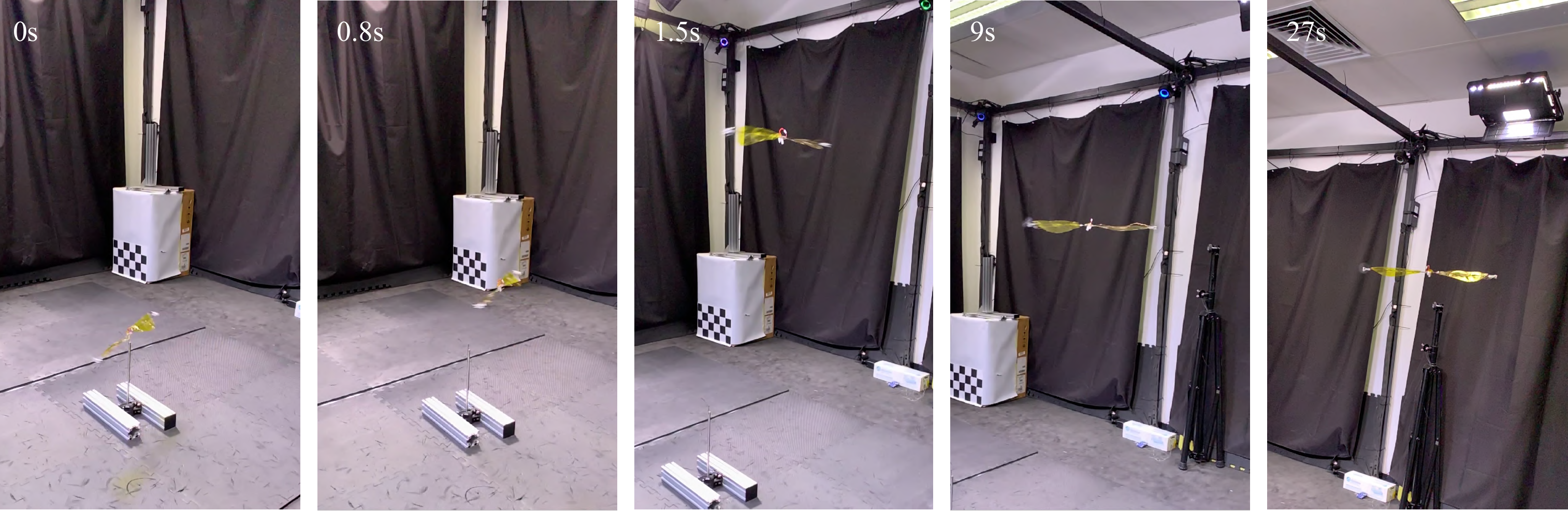}
\par\end{centering}
\caption{Frames at various times during the robot\textquoteright s flight.
The first three frames show the take off of the robot. The latter
two give its normal flight. \label{fig:PA}}
\vspace{-4mm}
\end{figure*}
The static tests suggest that the prototype with the optimal design
(robot \textcircled{\footnotesize{A}}) is capable of producing sufficient
thrust for flight. In the current prototypes, without an onboard flight
controller, we are unable to implement real-time feedback control.
Nevertheless, a revolving-wing robot possesses passive attitude stability
as long as certain conditions related to the inertial tensor are satisfied
\cite{ulrich2010falling}.

To demonstrate a flight, the robot was placed on a launch platform
that allows the robot to revolve and gain sufficient rotational speed
before lifting off. The robot was remotely controlled for propellers'
thrust as an indirect way to regulate its altitude manually by a human
operator. \Figref{PA} displays the video frames of the robot during
the takeoff and up to $27$ s after. At the beginning, the robot climbed
over $1.5$ m in less than two seconds. During the flight, we observed
minimal lateral drift. Correspondingly, the tip-path plane remained
relatively horizontal during the whole flight. \vspace{-1mm}

%% file: sec_conclusion.tex
\section{Conclusion}

Motivated by flight of a winged achene, in this work, we have proposed
a modeling framework for a motor-driven revolving-wing robot. Exploiting
the devised dynamic models, we fabricated a 13.8-gram robot with an
optimal design and demonstrated a hovering flight. The developed framework
entails uses of quasi-steady aerodynamic methods, namely momentum
theory and blade element method, for describing the dynamics of the
airfoils and propellers. The actuation system (motors and propellers)
was taken into account in the airfoil design in an attempt to maximum
the thrust generated by the robot. Based on the design optimization,
the manufactured robot has the maximum take-off weight of 310 mN and
thrust-to-weight ratio of 2.3. The revolving airfoils boost the thrust
by $\approx$50\% compared to multirotor designs. In a flight demonstration,
the robot displayed a stable hovering flight without active stabilization.

In the future, we plan to perform extensive measurements on both airfoils
and the actuation system for parameter identification in order to
improve the model accuracy. Further improvements can be achieved by
taking into consideration the aerodynamic interaction between the
airfoils and the propellers. To achieve position controlled flight,
flight dynamics must be extensively studied. \vspace{-1mm}